\documentclass[letterpaper]{article}
\usepackage{proceed2e}
\usepackage[margin=1in]{geometry}
\usepackage{graphicx}
\usepackage{times}
\usepackage{adjustbox}
\usepackage{amsmath}
\usepackage{amsthm}
\usepackage{amssymb}
\usepackage{graphicx}
\usepackage{natbib}
\title{Towards Adversarial Training with Moderate Performance Improvement for Neural Network Classification}

\author{
Xinhan Di$^1$, 
Pengqian Yu$^2$, 
Meng Tian$^2$, 
\\ 
$^1$ Huawei Technologies Noah's Ark Lab\\
$^2$ National University of Singapore\\
\\
dixinhan@huawei.com,
yupengqian@u.nus.edu,
tianmeng@u.nus.edu
}

\begin{document}

\maketitle

\begin{abstract}
It has been demonstrated that deep neural networks are prone to noisy examples particular adversarial samples during inference process. The gap between robust deep learning systems in real world applications and vulnerable neural networks is still large. Current adversarial training strategies improve the robustness against adversarial samples. However, these methods lead to accuracy reduction when the input examples are clean thus hinders the practicability. In this paper, we investigate an approach that protects the neural network classification from the adversarial samples and improves its accuracy when the input examples are clean.  We demonstrate the versatility and effectiveness of our proposed approach on a variety of different networks and datasets.
\end{abstract}

\section{Introduction}

Many high-performance deep learning applications in computer vision, speech recognition and other areas are susceptible to minimal changes of the inputs \cite{he2018decision}. Therefore, a robust system is required for the real-world applications when the input is affected by many interferences and noise. 

Current robust learning strategies in the context of neural network classification only play against adversarial samples, while the accuracy reduces when the samples are clean.  In this paper, we stand on the perspective of robust optimization and propose an approach to improve the performance of the networks on both clean and noise data. In particular, we propose a counterpart of the FSGM algorithm (\cite{goodfellow2014explaining}), which inherits the sign function for the robustness against permutation and helps the neural network leaves saddle points during the training process.  

\section{Related Work}

Adversarial training strategies are proposed for both attacks and defenses inspired by robust optimization. For example, a training procedure is provided that model parameters updates are augmented with the worst-case perturbations of the training data \cite{sinha2017certifiable}. A mixed strategy named stochastic activation pruning (SAP) is applied on the training of deep neural networks for the robustness against adversarial examples in \cite{he2018decision}. A potential application of local intrinsic dimensionality is studied to distinguish adversarial examples \cite{ma2018characterizing}. These adversarial training strategies are presented with better performance when the input examples are noisy. However, the above algorithms do not perform well when the input samples are free from noise. In the following, we introduce an algorithm that is capable to improve the performance without any assumption on the data quality.  

\section{Methodology}

In a standard deep learning task, we let $f$ denote the complex non-linear function of the deep neural network and  $\theta$ denote the parameters of $f$. Let $x$ and $y$ be the input and output sample of the network.  $y=f(x)$ denotes the standard function of a deep learning task. We use $sgn$ to denote the sign function and use $\delta (x)$ to define the dirac delta function. 
We use $\it{D}$ to represent the data distribution over pairs of examples $x\in \mathbb{R}^{d_1}$ and labels $y \in \mathbb{R}^{d_2}$. Denote $L(\theta, x, y)$ as the loss function, and the goal is to find model parameters $\theta$ that minimize the risk $\mathbb{E}_{(x,y)}[L(\theta, x, y)]$. This empirical risk minimization (ERM) is successful for finding classifiers with small population task. However, it is not robust enough against some common disturbances. In particular, the model incorrectly classifies $\hat{x}$ as belonging to a different class of $x$ if $\hat{x}$ is very close to $\hat{x}$.

In order to impose robustness, the population risk $\mathbb{E}_{\it{D}}[L]$ is formulated as the following
\begin{equation}
    \rho(\theta) = \mathbb{E}_{(x,y) \sim \it{D}} \left[\max_{\delta \in \it{S}} L(\theta, x + \delta, y)\right]. \\
\end{equation}
The objective is to minimize the above population risk 
$$\min{\rho(\theta)}.$$

We next define $g(x)$ to be the function which produces adversarial samples from clean samples. That is, $\hat{x} = g(x)$. The population risk $\mathbb{E}_{\it{D}}[L]$ is represented as $\rho(\theta) = \mathbb{E}_{(x,y) \sim \it{D}} \left[\max_{\delta \in \it{S}} L(\theta, g(x), y)\right]$. The corresponding function is denoted as $y=f(g(x))$. In the following, we first propose an algorithm, and analyzed the Jacobian of the deep neural network when the algorithm is applied. We further analyze the Hessian. Based on the analysis, we could safely say that the proposed algorithm dose not introduce extra saddle points and may be very likely help the neural network leave saddle points on a sharper direction of the loss function surface. 

\subsection{Analysis of Jacobian}
We propose the following algorithm: 
$$g(x)=x + \epsilon xsgn(x).$$  

We have 
$$\frac{\partial g(x)}{\partial x} = 1 + \epsilon (sgn(x) + x \delta'),$$

\begin{equation*}
\begin{aligned}
\frac{\partial^{2} g(x)}{\partial x^2} &= \epsilon (\delta(x) + \delta(x) + x \delta'(x))\\
& = \epsilon (\delta(x) + \delta(x) - \delta(x)) \\&= \epsilon \delta (x)
\end{aligned}
\end{equation*}

Firstly, the Jacobian is given as following.
\begin{equation*}
\begin{aligned}
J(\theta) &= \frac{\partial f(g(x))}{\partial x}\\
&= \frac{\partial f(x)}{\partial x}(1 + \epsilon (sgn(x) + x \delta'))\\
&= \frac{\partial f(x)}{\partial x}(1 + \epsilon sgn(x))
\end{aligned}
\end{equation*}
If $\epsilon$ is smaller than $1$, no extra extreme points will be introduced. The extreme points are all from $\frac{\partial f(x)}{\partial x} = 0$. 

\subsection{Analysis of Hessian}

The Hessian is given as following.
\begin{equation*}
\begin{aligned}
H(\theta)=& \frac{\partial^{2} f(g(x))}{ \partial x^2}\\
= &\frac{\partial^{2} f(x)}{\partial x^2} \frac{\partial g(x)}{\partial x} \frac{\partial g(x)}{\partial x} + \frac{\partial f(x)}{\partial x} \frac{\partial^{2} g(x)}{\partial x^{2}}\\
=& \frac{\partial^{2} f(x)}{\partial x^2} (1 + \epsilon (sgn(x) + x \delta(x)))^{2} + \frac{\partial f(x)}{\partial x}(\epsilon \delta(x))
\end{aligned}
\end{equation*}

For any extreme points such that $\frac{\partial f(x)}{\partial x}=0$, the Hessian equation is given by
$$\newline H(\theta)= \frac{\partial^{2} f(x)}{\partial x^2}(1 + \epsilon^2 sgn(x)sgn(x)).$$

It could be observed that the change rate of the Jacobian is increased as $1 + \epsilon^2 sgn(x)sgn(x)\geq1$. This implies a higher rate of change which helps the neural networks leave saddle points during the training process.  

\subsection{Implementation}
The proposed $g(x)$ could be applied in the input space and the hidden space of the deep neural networks. That is, $x$ could be samples of the input and the vectors in the hidden space of the deep neural networks.

In order to apply the adversarial training strategy on-line without extra computational burden, the proposed algorithm dose not rely on the calculation of gradients like FGSM (\cite{goodfellow2014explaining}). Instead, the perturbation item is calculated feed-forward once the samples in the input/hidden space are calculated. 

Let $x$ denote the samples in the input/hidden space. In the following experiments, $g(x)$ has the form
\begin{equation*}
g(x) = x + \epsilon x \odot l(x).
\end{equation*}
Here
\begin{equation*}
\left\{  
    \begin{array}{lr}  
    l(x) = sgn(x) \quad w.p.\quad p ,&\\ 
    l(x) = -sgn(x) \quad w.p.\quad 1-p.\\  
    \end{array}  
\right.  
\end{equation*}

\section{Numerical experiments}

\begin{figure*}[h!]
\begin{center}
   \includegraphics[width=0.8\linewidth]{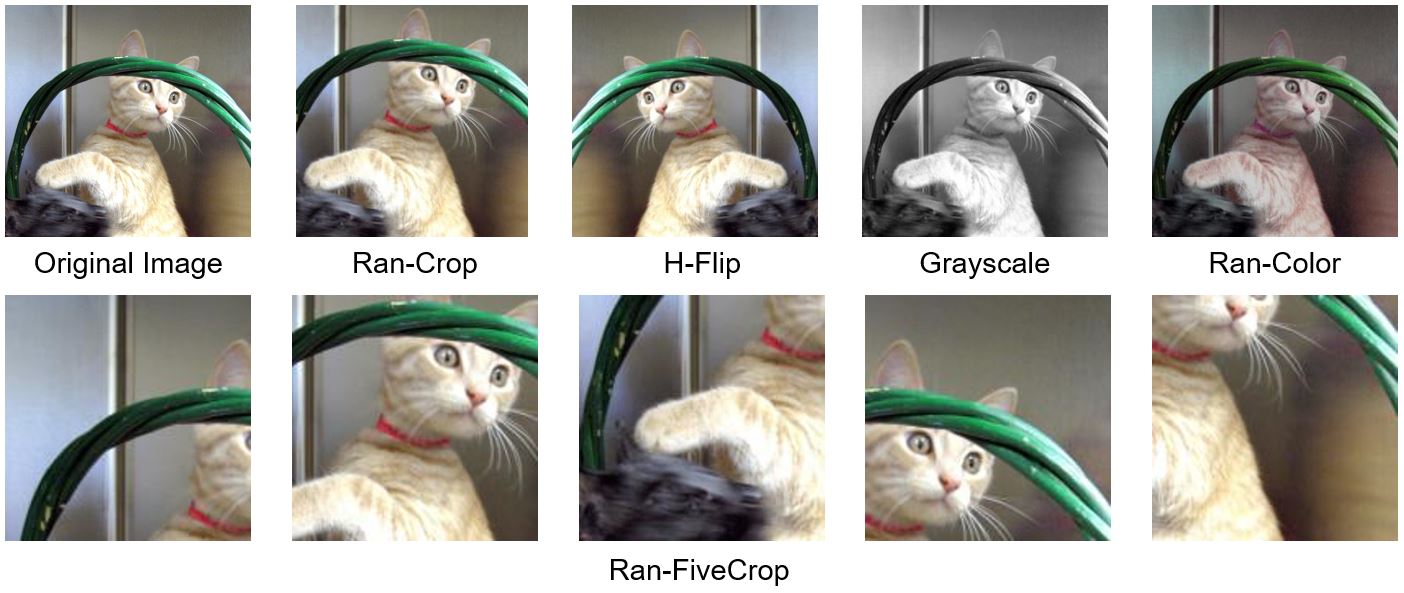}
\end{center}
   \caption{Five common attacks.}
\label{fig:1}
\end{figure*}

\begin{table*}[h!]
\caption{Top-1 classification accuracy on ImageNet-1k for five different disturbances in the input space.}
\begin{center}
\begin{adjustbox}{width=0.85\textwidth}
\begin{tabular}{|l|c|c|c|c|c|}
\hline
\textbf{Model} & \textbf{Ran-Crop} & \textbf{Ran-Hlip} & \textbf{Ran-GrayScale}(\%) & \textbf{Ran-Color}(\%) & \textbf{Random Five Crop}(\%)\\
\hline
\hline
\multicolumn{6}{|c|}{Imagenet} \\
\hline
FGSM-CondenseNet(G=C=8) & 68.27 & 69.42 & 68.03 & 69.21 & 58.17\\
CondenseNet(G=C=8) & $70.04\pm0.05$& $71.00\pm0.03$ & $69.79\pm0.04$ & $71.05\pm0.04$ & $60.03\pm0.05$\\
R-CondenseNet(G=C=8) & $\textbf{70.38}\pm{0.03}$ & $\textbf{71.29}\pm{0.05}$ & $\textbf{70.06}\pm{0.03}$ & $\textbf{71.25}\pm{0.02}$ & $\textbf{60.45}\pm{0.04}$\\
FGSM-CondenseNet(G=C=4) & 70.34 & 71.48 & 70.72 & 71.39 & 60.04\\
CondenseNet(G=C=4) & 7$2.92\pm{0.02}$ & $73.78\pm{0.05}$ & $72.74\pm{0.03}$ & $73.76\pm{0.03}$ & $62.52\pm{0.04}$\\
R-CondenseNet(G=C=4) & $\textbf{73.34}\pm0.03$ & $\textbf{74.12}\pm0.02$ & $\textbf{73.05}\pm0.05$ & $\textbf{74.07}\pm0.06$ & $\textbf{63.26}\pm0.04$\\
\hline
\end{tabular}
\end{adjustbox}
\end{center}
\label{table1}
\end{table*}

\begin{table*}[h!]
\caption{Top-1 and Top-5 classification accuracies of state-of-art regular models on ImageNet-1k.}
\begin{center}
\begin{adjustbox}{width=0.65\textwidth}
\begin{tabular}{|l|c|c|c|c|c|c|}
\hline
\textbf{Model} & \textbf{Image Size} & \textbf{Params} & \textbf{Mul-Adds} & \textbf{Tops-1}(\%) & \textbf{Tops-5}(\%)\\
\hline
\hline
\multicolumn{6}{|c|}{Imagenet} \\
\hline
\hline
SENet & $320 \times 320$ & 145.8M & 42.4B & 82.7 & 96.2 \\
NASNet-A  & $331 \times 331$ & 88.9M & 23.8B & 82.7 & 96.2 \\
FGSM-NASNet-A  & $331 \times 331$ & 88.9M & 23.8B & 80.6 & 92.1 \\
R-NASNet-A & $331 \times 331$ & 88.9M & 23.8B &$\textbf{83.0}\pm\textbf{0.03}$ & $\textbf{96.5}\pm\textbf{0.05}$\\
\hline
\end{tabular}
\end{adjustbox}
\end{center}
\label{table2}
\end{table*}

\begin{table*}[h!]
\caption{Top-1 and Top-5 classification error rate of state-of-art compact models on ImageNet-1k.}
\begin{center}
\begin{adjustbox}{width=0.65\textwidth}
\begin{tabular}{|l|c|c|c|c|c|c|}
\hline
\textbf{Model} & \textbf{FLOPs} & \textbf{Params} & \textbf{Tops-1}(\%) & \textbf{Tops-5}(\%)\\
\hline
\multicolumn{5}{|c|}{Imagenet} \\
\hline
Inception V1 & 1,448M & 6.6M & 30.2 & 10.1 \\
1.0 MobileNet-224 & 569M & 4.2M & 29.4 & 10.5 \\
ShffleNet 2x & 524M & 5.3M & 29.1 & 10.5 \\
NASNet-B(N=4) & 488M & 5.3M & 27.2 & 9.0 \\
NASNet-C(N=3) & 558M & 4.9M & 27.5 & 9.0 \\
CondenseNet(G=C=8) & 274M & 2.9M & 29.0 & 10.0 \\
CondenseNet(G=C=4) & 274M & 2.9M & 26.2 & 8.3 \\
FGSM-CondenseNet(G=C=8) & 274M & 2.9M & 30.8 & 12.3 \\
FGSM-CondenseNet(G=C=4) & 529M & 4.8M & 27.9 & 10.7 \\
\hline
R-CondenseNet(G=C=8)  & 274M & 2.9M &$\textbf{28.6}\pm \textbf{0.05}$ &$\textbf{9.9}\pm \textbf{0.06}$ \\
R-CondenseNet(G=C=4)  & 529M & 4.8M &$\textbf{25.8}\pm \textbf{0.04}$ &$\textbf{8.1}\pm \textbf{0.05}$ \\
\hline
\end{tabular}
\end{adjustbox}
\end{center}
\label{table3}
\end{table*}

\begin{table*}[h!]
\caption{Middle-scaled supervised classification: MNIST and SVHN.}
\begin{center}
\begin{adjustbox}{width=0.85\textwidth}
\begin{tabular}{|l|c|c|c|c|c|c|}
\hline
Model & Depth & Factor & Dropout & Standard Aug & Random Erase & Error(\%)\\
\hline
\hline
\multicolumn{7}{|c|}{MNIST} \\
\hline
LeNet\cite{lecun1998gradient} & - & - & $ \surd$ & - & - & 0.50\\
FGSM-LeNet & - & - & $ \surd$ & - & - & 0.60\\
R-LeNet& - & - & $ \surd$ & - & - & \textbf{0.28}\\
CapsNet\cite{sabour2017dynamic} & - & - & $\surd$ & - & - &  0.35\\
FGSM-CapsNet & - & - & $ \surd$ & - & - & 0.51\\
R-CapsNet & - & - & $\surd$ & - & - & \textbf{0.30}\\
\hline
\multicolumn{7}{|c|}{SVHN} \\
\hline
S-ResNet \cite{huang2016deep} & 110 & - & $\surd$ & $\surd$ & - & 1.75\\
FGSM-S-ResNet & 110 & - & $\surd$ & $\surd$ & - & 1.90\\
R-S-ResNet & 110 & - & $\surd$ & $\surd$ & - & \textbf{1.70}\\
W-ResNet \cite{BMVC2016_87}& 16 & k = 8 & $\surd$ & $\surd$ & - & 1.54\\
FGSM-S-Resnet & 16 & k = 8 & $\surd$ & $\surd$ & - & 1.71\\
R-W-ResNet & 16 & k = 8 & $\surd$ & $\surd$ & - &\textbf{1.49}\\
\hline
\end{tabular}
\end{adjustbox}
\end{center}
\label{table4}
\end{table*}

\begin{table*}
\caption{Middle-scaled supervised classification: CIFAR10 and CIFAR100.}
\begin{center}
\begin{adjustbox}{width=0.85\textwidth}
\begin{tabular}{|l|c|c|c|c|c|c|}
\hline
Model & Depth & Factor & Dropout & Standard Aug & Random Erase & Error(\%)\\
\hline
\multicolumn{7}{|c|}{CIFAR10}\\
\hline
W-ResNet \cite{BMVC2016_87}& 28 & k = 10 & $\surd$ & $\surd$ & $\surd$ & 3.1\\
FGSM-W-ResNet & 28 & k = 10 & $\surd$ & $\surd$ & $\surd$ & 4.2\\
R-W-ResNet & 28 & k = 10 & $\surd$ & $\surd$ & $\surd$ &\textbf{2.7}\\
\hline
\multicolumn{7}{|c|}{CIFAR100} \\
\hline
InceptionV3 \cite{szegedy2015going} & 48 & - & $\surd$ & $\surd$ & - & 22.69\\
FGSM-InceptionV3  & 48 & - & $\surd$ & $\surd$ & - & 23.41\\
R-InceptionV3 & 48 & - & $\surd$ & $\surd$ & - & \textbf{21.80}\\
W-ResNet \cite{BMVC2016_87} & 28 & k = 10 & $\surd$ & $\surd$ & $\surd$ & 17.73\\
FGSM-W-ResNet & 28 & k = 10 & $\surd$ & $\surd$ & $\surd$ & 18.61\\
R-W-ResNet & 28 & k = 10 & $\surd$ & $\surd$ & $\surd$ &\textbf{17.18}\\
\hline
\hline
\end{tabular}
\end{adjustbox}
\end{center}
\label{table5}
\end{table*}

We evaluate the proposed form of robust algorithm $g(x)$ where $p=0.5$. The evaluation is made on a variety of popular datasets, including small-scale, middle-scale and large-scale datasets (MNIST \cite{lecun1998gradient}, CIFAR10, CIFAR100 \cite{krizhevsky2009learning}, SVHN \cite{netzer2011reading} and Imagenet-1k \cite{russakovsky2015imagenet}). Extensive experimental evaluations are presented in two aspects including performance improvement for the clean samples in input space and the noisy samples in input space (see Figure \ref{fig:1}). We evaluate the FGSM for base models and the proposed algorithm for based models, called R-base models. As shown in Table \ref{table1}, the accuracy is improved when the inputs are noisy compared with base models and FGSM for base models. As shown in Table \ref{table2} to Table \ref{table5}, the accuracy is improved when the inputs are clean. 

\section{Discussion}
We propose an approach that help neural networks achieve both robustness against noisy inputs and higher accuracy for clean input. It enhances the practicality of neural networks such that the input can be clean or noisy.

This is an initial work that only five common types of noise are evaluated. In the real world, the types of noise are unknown and it remains unclear whether the inputs is attacked by the disturbance or not. In addition, the base model is also a black box that the gradients are hard to obtain. A systematic way of overcoming these problems deserve one's attention.

\bibliographystyle{named}
\bibliography{ijcai18}

\end{document}